\documentclass[10pt,twocolumn,letterpaper]{article}

\usepackage[pagenumbers]{cvpr} 

\usepackage{graphicx}
\usepackage{amsmath}
\usepackage{amssymb}
\usepackage{booktabs}
\usepackage{makecell}
\usepackage{multirow}
\usepackage[pagebackref,breaklinks,colorlinks]{hyperref}

\usepackage[capitalize]{cleveref}
\crefname{section}{Sec.}{Secs.}
\Crefname{section}{Section}{Sections}
\Crefname{table}{Table}{Tables}
\crefname{table}{Tab.}{Tabs.}
\newtheorem{definition}{Definition}
\newtheorem{proposition}{Proposition}

\def\cvprPaperID{28} 
\def\confName{CVPR Workshop}
\def\confYear{2022}

\begin{document}

\title{Alleviating Representational Shift for Continual Fine-tuning}

\author{Shibo Jie\qquad Zhi-Hong Deng\thanks{Corresponding author}\qquad Ziheng Li\\
School of Artificial Intelligence, Peking University\\
{\tt\small \{parsley, zhdeng, liziheng\}@pku.edu.cn}
}
\maketitle

\begin{abstract}
We study a practical setting of continual learning:  fine-tuning on a pre-trained model continually. Previous work has found that, when training on new tasks, the features (penultimate layer representations) of previous data will change, called representational shift. Besides the shift of features, we reveal that the intermediate layers' representational shift (IRS) also matters since it disrupts batch normalization, which is another crucial cause of catastrophic forgetting. Motivated by this, we propose ConFiT, a fine-tuning method incorporating two components, cross-convolution batch normalization (Xconv BN) and hierarchical fine-tuning. Xconv BN maintains pre-convolution  running  means  instead of post-convolution,  and recovers post-convolution ones before testing, which corrects the inaccurate estimates of means under IRS. Hierarchical fine-tuning leverages a multi-stage strategy to fine-tune the pre-trained network, preventing massive changes in Conv layers and thus alleviating IRS. Experimental results on four datasets show that our method remarkably outperforms several state-of-the-art methods with lower storage overhead. Code: \url{http://github.com/JieShibo/ConFiT}.
\end{abstract}

\section{Introduction}

An \emph{artificial neural network} (ANN) well-trained on a task suffers from \emph{catastrophic forgetting} when learning a new task: its performance on already learned tasks drops dramatically after learning a new one \cite{1989cataf,14r}. However, in a real-world scenario where re-training on past data may be expensive or infeasible, we expect ANNs to be able to learn continually, acquiring new knowledge without interfering with previously learned skills. This learning paradigm is referred to as \emph{continual learning}.
\begin{figure}[t]
\centering
\includegraphics[height=0.26\textwidth]{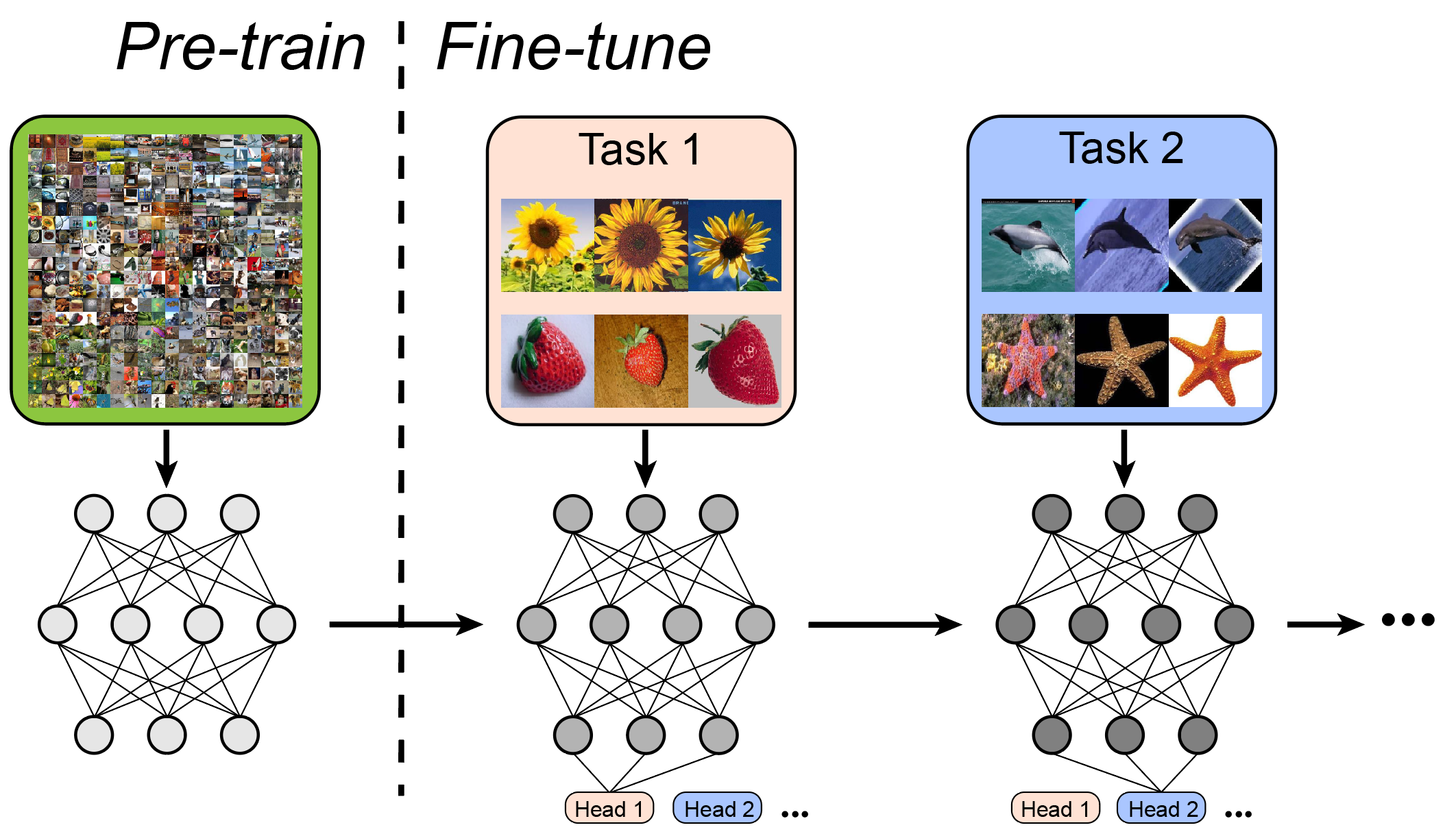}
\caption{Continual Fine-tuning on a pre-trained model in multi-head (task-incremental) scenario.}
\label{fig1.5}
\end{figure}

Meanwhile,  \emph{pre-trained models} have already shown overwhelming results in many fields of deep learning, such as \emph{computer vision} and \emph{neural language processing} (NLP) \cite{bert,gpt,ptcv}. By pre-training an ANN on a large-scale dataset and then fine-tuning all or part of the network parameters on the downstream dataset, the general knowledge obtained from large-scale datasets can be transferred to downstream tasks. 
Due to the effectiveness of pre-training \& fine-tuning paradigm, using a pre-trained model as initialization has already been adopted as a generic approach in continual learning of NLP \cite{survey2,nlp1,nlp2,nlp3}. 

Whereas in vision, most previous works of continual learning prefer to initialize the network randomly. Other works either directly leverage the pre-trained model as a frozen feature extractor, upon which shallower classifiers or task-specific masks are continually learned \cite{cvpr2,fearnet,piggyback}; or investigate the performance of their methods when fine-tuning continually on hard fine-grained datasets \cite{afec,cpr,nccl,mas,sdc} and/or tough scenarios \cite{sdc,cvprw1,nccl}. 
Since our ultimate aim is to deploy a high-performance continual learning system in the real world, it is sensible to leverage the pre-trained models and their general knowledge. Therefore, how to fine-tune a network without catastrophic forgetting given its pre-trained parameters is a key issue for designing effective methods in continual learning.

So, what is the crucial cause of catastrophic forgetting when fine-tuning continually? We here follow the multi-head setting adopted by some previous works \cite{ccll,cpr,ewc}, under which each task monopolizes a specific output fully-connected (FC) layer and the other layers are shared among tasks, as shown in \cref{fig1.5}. Ideally, the penultimate layer representations of previous tasks should be kept unchanged when learning new tasks. Otherwise, the distorted representations may not correspond with the output layer, because the output layers of previous tasks are not updated. We refer to this phenomenon as \emph{representational shift}. The shift of the penultimate layer representations has already been discussed by \cite{sdc,lucas1,lucas2}, but we will step further here: the representational shift of other intermediate layers in the network is also noteworthy, since it significantly disrupts \emph{Batch Normalization} (BN) \cite{bn}. 
\begin{figure}[t]
\centering
\includegraphics[height=0.10\textwidth]{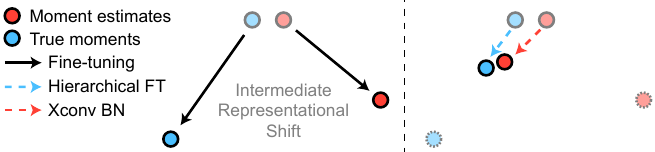}
\caption{Fine-tuning (left) and ConFiT (right). After fine-tuning on new tasks, the true moments of previous tasks' intermediate representations are shifted, and the running moment estimates of BN are biased towards new tasks. ConFiT alleviates IRS via hierarchical fine-tuning and corrects running means via Xconv BN.}
\label{fig1}
\end{figure}

BN is a widely used technique in deep backbone networks such as ResNet \cite{resnet}. While testing, BN normalizes the intermediate representations using running estimates (means and variances) of the global moments calculated during training. In continual learning, two problems arise when we directly employ BN in networks. First, the running moments will be inevitably biased towards new tasks. Second, even if we keep the running moments unchanged, the representational shift of intermediate layers (i.e., \emph{intermediate representational shift}, IRS) still leads to changes in the distribution of intermediate representations. Due to these two problems, the BN becomes uncontrollable in continual learning, and thus we cannot normalize intermediate representations accurately when testing on previous tasks. As a result, the model's performance on previous tasks is degraded, which causes catastrophic forgetting.

In this paper, we propose \emph{ConFiT}, a method to address these issues on both sides via two novel techniques, \emph{cross-convolution batch normalization (Xconv BN)} and \emph{hierarchical fine-tuning}, as illustrated in \cref{fig1}. Xconv BN is designed on the basis of BN, but can provide more accurate running estimates under IRS. For BNs placed after convolutional (Conv) layers, we calculate the pre-convolution running means before the Conv layers during training, and use them to recover the post-convolution running means. Hierarchical fine-tuning prevents massive changes in Conv layers by fine-tuning the network with a multi-stage strategy and thus alleviates IRS.

In summary, our contributions are as follows:

\begin{itemize}
    \item We reveal that intermediate representational shift (IRS) deserves to be noticed since it disrupts BN and consequently leads to forgetting. As far as we know, no previous work has presented this issue.
    \item We propose ConFiT, a novel fine-tuning method that both alleviates IRS and corrects BN estimates. 
    \item Experimental results show that the proposed method remarkably promotes the performance of fine-tuning and outperforms several state-of-the-art baselines with lower additional storage overhead.
\end{itemize}

\section{Related Work}

\subsection{Continual Learning}
Continual learning methods focus on alleviating catastrophic forgetting, which can be categorized into three groups \cite{survey1}. \emph{Regularization-based} methods  measure the importance of parameters in previous tasks, and then impose a regularization term on the loss to prevent important parameters from significant deviation \cite{ewc,si,rwalk,mas,cpr,afec}. \emph{Replay-based} methods maintain a limited memory space to store samples from previous tasks, and jointly train them with new data to overcome forgetting \cite{er,icarl,gss,mir,bic,gdumb,cpe,der}. \emph{Parameter-isolation} methods offer distinct parameters for each task, so as to isolate interference among tasks \cite{rps,ccll,piggyback,pathnet}.

Pre-trained models such as BERT \cite{bert} and GPT2 \cite{gpt} have already been used in NLP tasks of continual learning  \cite{nlp1,nlp2,nlp3}. However, in vision tasks, most recent works prefer to start from random initialization, or use a frozen pre-trained model as a feature extractor, which simplifies the problem by only learning the shallow classifier continually. There are only a few works that have made limited attempts to fine-tune pre-trained vision models. For instance, \cite{cpr,afec,mas} investigate the fine-tuning performance of their methods on hard datasets (e.g., CUB200 \cite{cub} and CORe50 \cite{core50}), and \cite{nccl,sdc,cvprw1} use pre-trained initialization under a more difficult class-incremental setting where new classes and patterns emerge over time. Whereas in this paper, we directly study the forgetting on pre-trained model and suggest strategies for fine-tuning continually.

\subsection{Batch Normalization}
Batch normalization (BN) is widely used in deep neural networks to accelerate training and reduce the sensitivity to initialization \cite{bn}. To be specific, in a convolutional neural network, the activation $\boldsymbol{a}$ has a shape of $(B, C, H, W)$, standing for batch size, channel size, height and width, respectively. During training, BN calculates moments as:
\begin{align}
\boldsymbol{\mu}_{BN} = \frac{1}{BHW}\sum_{b=1}^B\sum_{h=1}^H\sum_{w=1}^W\boldsymbol{a}_{b,:,h,w}
\end{align}
\begin{align}
\boldsymbol{\sigma}^2_{BN} = \frac{1}{BHW}\sum_{b=1}^B\sum_{h=1}^H\sum_{w=1}^W\left(\boldsymbol{a}_{b,:,h,w}-\boldsymbol{\mu}_{BN}\right)^2
\end{align}

where $\boldsymbol{\mu}_{BN},\boldsymbol{\sigma}^2_{BN}\in \mathbb{R}^C$. Meanwhile, BN maintains momentum-updated running moments to approximate global moments: 
\begin{align}\boldsymbol{\mu}_{r} \leftarrow \boldsymbol{\mu}_{r} + \eta \left(\boldsymbol{\mu}_{BN}-\boldsymbol{\mu}_{r}\right), \quad\boldsymbol{\sigma}^2_{r} \leftarrow \boldsymbol{\sigma}^2_{r} + \eta \left(\boldsymbol{\sigma}^2_{BN}-\boldsymbol{\sigma}^2_{r}\right)\end{align}

The normalized activation is: 
\begin{align}\boldsymbol{a'} = \boldsymbol{\gamma}\left(\frac{\boldsymbol{a} - \boldsymbol{\mu}}{\sqrt{\boldsymbol{\sigma}^2 + \epsilon}}\right)+\boldsymbol{\beta}\end{align}
where $\boldsymbol{\mu}=\boldsymbol{\mu}_{BN}, \boldsymbol{\sigma}=\boldsymbol{\sigma}_{BN}$ in training, and $\boldsymbol{\mu}=\boldsymbol{\mu}_{r}, \boldsymbol{\sigma}=\boldsymbol{\sigma}_{r}$ in testing. $\boldsymbol{\gamma},\boldsymbol{\beta}\in \mathbb{R}^C$ are trainable affine parameters. All operations are broadcast alone the axes of $B$, $H$ and $W$.

An obvious weakness of BN is that the training may fail due to unstable $\boldsymbol{\mu}_{BN}$ and $\boldsymbol{\sigma}^2_{BN}$ when data batches are small and/or non-i.i.d., especially when the model is confronted with a continual unstable data stream. \cite{cvprw1} suggests replacing BN with Batch Renormalization (BRN) \cite{brn} so as to bypass this drawback. But this weakness will not be exposed in task-incremental learning where the data batches are i.i.d. within each single task.

Another flaw of BN is that the moments used in testing are estimated with training data, which may lead to trouble when the estimates do not match the true distribution of testing data. This inconsistency becomes more severe in continual learning, which we will discuss comprehensive in \cref{s3.1}. For this reason, a novel normalization method has been proposed by a recent study \cite{cn} to replace BN in online continual learning. In this paper, we will improve BN in another way to address this issue, which only differs from BN in testing without modifying its normalization behavior in training.

\section{Methodology}
\subsection{Preliminaries \& Motivation}
\label{s3.1}
In this paper, we consider a fundamental setting where task boundaries are available. The model is supposed to learn a sequence of tasks: $\mathcal{T} = \{\mathcal{T}_1, \mathcal{T}_2,..., \mathcal{T}_T\}$. At each step $t$, the model can only have access to training set of $\mathcal{T}_t = \{(\boldsymbol{x}_i^{tr}, y_i^{tr}, t)\}$. After training, the model will be tested on test set of all learned tasks $\{(\boldsymbol{x}_i^{te}, y_i^{te}, j)|j=1,...,T\}$. Following multi-head setting in previous works, we denote the feature extractor of the model as  a mapping $f$, and the classifier head of task $\mathcal{T}_t$ as another mapping $h_t$. Then at step $t$, without any continual learning approaches, we train the network by minimizing the empirical risk $\mathcal{L}_t = \frac{1}{|\mathcal{T}_t|}\sum_{i=1}^{|\mathcal{T}_t|}\mathcal{L}(h_t\circ f(\boldsymbol{x}^{tr}_i), y^{tr}_i)$.

As the training goes on, $f$ is evolving all the time, whereas $h_t$ is updated only at step $t$. This is to say that at $t'>t$, the representations outputted by $f$ have changed and may no longer correspond with the classifier $h_t$. We refer to this phenomenon as representational shift of the penultimate layer. 

To generalize this phenomenon to other layers, we now consider a more complicated case: a network with BN. We decompose the feature extractor $f$ into $f_{BN}\circ f_1$. Since $f_{BN}$ maintains running moments for testing, at $t'>t$, the running moments will be flooded by moments of the newest task, and thus the intermediate $f_1(\boldsymbol{x})$ for $\boldsymbol{x}\in \mathcal{T}_t$ will not be normalized correctly during testing.

A trivial solution to this issue is to store specific moments for each task respectively. However, as discussed above, even if the running moments of $f_{BN}$ are task-specific, the $f_1$ is still evolving all the time, which also results in IRS --- the distribution of intermediate representation $f_1(\boldsymbol{x})$ for $\boldsymbol{x}\in \mathcal{T}_t$ has changed, so that the running moments of previous task $\mathcal{T}_t$ have been no longer representative of this distribution. This IRS intensifies the inconsistency between training and testing of BN, and thus leads to catastrophic forgetting.

We now introduce the two components of ConFiT.

\subsection{Cross-convolution Batch Normalization}
\label{s3.2}
Since BN usually follows a Conv layer, we now consider an intermediate fragment of network composed of a Conv layer and a BN layer: $f_{BN}\circ f_{Conv}$. Firstly, for $\boldsymbol{\mu}_{r}$, $\boldsymbol{\sigma}_{r}$, $\boldsymbol{\beta}$, and $\boldsymbol{\gamma}$ of $f_{BN}$, we store specific ones for each task to prevent them from leaning to new tasks. For a previous task, if input activation $\boldsymbol{a}$ of this fragment is fixed, then the IRS disrupting $f_{BN}$ is caused by the change of $f_{Conv}$. We start our analysis with an interesting relation between the input's and output's mean of a Conv layer. For convenience, we first declare two following operators:

\begin{definition}[Average Pooling]
Suppose activation $\boldsymbol{a}$ has a shape of $(B, C, H, W)$, then $AvgPool(\boldsymbol{a})$ has a shape of $(C)$, in which
\begin{align}AvgPool(\boldsymbol{a})_{c}=\frac{1}{BHW}\sum_{b'=1}^B\sum_{h'=1}^H\sum_{w'=1}^W\boldsymbol{a}_{b',c,h',w'}\end{align}
\end{definition}

\begin{definition}[Dimension-Preserving Average Pooling]
Suppose activation $\boldsymbol{a}$ has a shape of $(B, C, H, W)$, then $AvgPool_{DP}(\boldsymbol{a})$ also has a shape of $(B, C, H, W)$, in which
\begin{align}AvgPool_{DP}(\boldsymbol{a})_{b,c,h,w}=AvgPool(\boldsymbol{a})_{c}\end{align}
\end{definition}
$AvgPool_{DP}$ is dimension-preserving because it does not change the shape of input as $AvgPool$ does. Note that if we broadcast $AvgPool(\boldsymbol{a})$ along the axes of $B$, $H$ and $W$, we will get $AvgPool_{DP}(\boldsymbol{a})$. With these operators, we show an invariance of the Conv layer's output:
\begin{proposition} Suppose $f_{Conv}(\cdot)$ denotes a 2D-Conv layer with $stride=1$ and $padding=kernel\_size-1$. Then,
\begin{align}
AvgPool\left(f_{Conv}(\boldsymbol{a})\right)=
AvgPool(f_{Conv}(AvgPool_{DP}(\boldsymbol{a}))
\end{align}
\end{proposition}
Proof and other variants when $stride\neq1$ or $padding\neq kernel\_size-1$ are in Appendix. 

The $f_{BN}$ following $f_{Conv}$ calculates mean as:
\begin{align}\boldsymbol{\mu}_{BN} = AvgPool(f_{Conv}(\boldsymbol{a}))\end{align}
From Prop. 1 we know that this mean can also be calculated as:
\begin{align}\boldsymbol{\mu}_{BN} = AvgPool(f_{Conv}(AvgPool_{DP}(\boldsymbol{a}))\end{align}

This is to say that, if we assume the distribution of input $\boldsymbol{a}$ is certain, we can always get the accurate post-convolution $\boldsymbol{\mu}_{BN}$ in testing by only storing its pre-convolution mean $AvgPool(\boldsymbol{a})$ during training, no matter how the weight of $f_{Conv}$ changes later. However, if we directly store $\boldsymbol{\mu}_{BN}$ instead (or use moving average $\boldsymbol{\mu}_{r}$ to approximate global $\boldsymbol{\mu}_{BN}$ when batch learning) as BN does, the changed Conv layer will make the stored training $\boldsymbol{\mu}_{BN}$ (or $\boldsymbol{\mu}_{r}$) unequal to true testing $\boldsymbol{\mu}_{BN}$. 
\begin{figure}[t]
\centering
\includegraphics[height=0.25\textwidth]{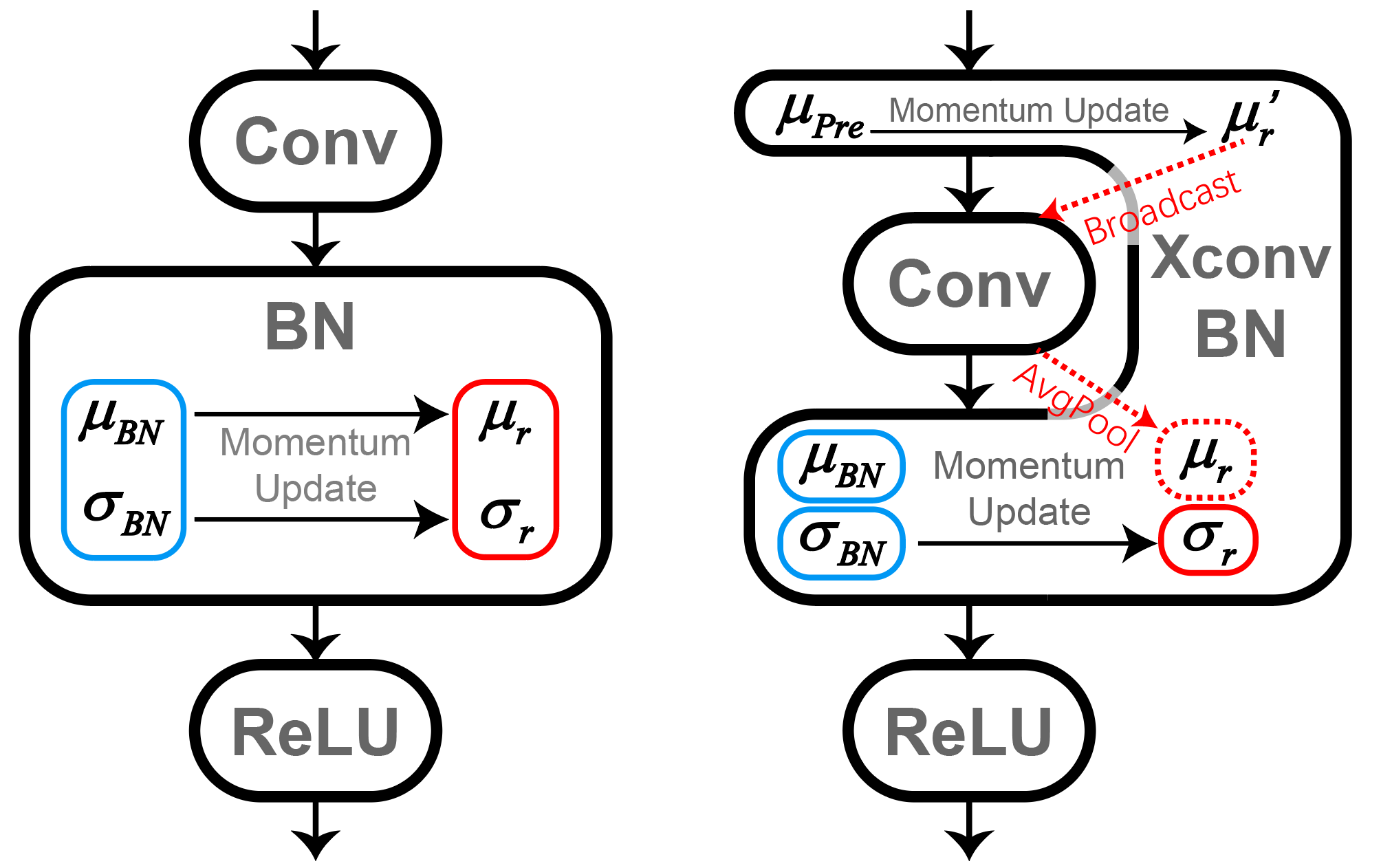}
\caption{BN (left) and Xconv BN (right). Moments in blue and red circles are used for training and testing, respectively. $\boldsymbol{\mu}_r$ of BN is inaccurate since Conv layer has changed, so Xconv BN calculates the post-convolution $\boldsymbol{\mu}_r$ with stable pre-convolution $\boldsymbol{\mu'}_{r}$ before testing.}
\label{fig4.1}
\end{figure}

On the other hand, if $f_{BN}$ normalizes the mean accurately, its outputs have a certain mean vector --- equal to its $\boldsymbol{\beta}$ parameter. So the pre-convolution mean of the next Conv layer is also certain --- exactly this $\boldsymbol{\beta}$ (if there are no other layers between them). Then as our analysis above, the pre-convolution mean of the next Conv layer will also be certain. Inductively, all BN layers will normalize the means accurately when testing. 

But in practice, there are also nonlinear activation functions between a Conv layer and the BN layer ahead of it, so we cannot directly use $\boldsymbol{\beta}$ of the previous BN as the pre-convolution mean. Instead, the proposed Xconv BN maintains a momentum-updated pre-convolution running mean $\boldsymbol{\mu'}_{r}$ instead of $f_{BN}$'s $\boldsymbol{\mu}_{r}$. $\boldsymbol{\mu'}_{r}$ is updated using batch pre-convolution means $\boldsymbol{\mu}_{Pre}=AvgPool(\boldsymbol{a})$:
\begin{align}\boldsymbol{\mu'}_{r} \leftarrow \boldsymbol{\mu'}_{r} + \eta \left(\boldsymbol{\mu}_{Pre}-\boldsymbol{\mu'}_{r}\right)\end{align}
Before testing, we uses it to calculate the approximate post-convolution global mean of this Conv layer:
\begin{align}\boldsymbol{\mu}_{r} = AvgPool(f_{Conv}(Broadcast(\boldsymbol{\mu'}_{r})))\end{align}

As for $\boldsymbol{\sigma}_{r}$, we will empirically prove that it is not necessary to correct it in \cref{s4.2}, so we here retain the post-convolution $\boldsymbol{\sigma}_{r}$ of BN in Xconv BN. The post-convolution $\boldsymbol{\sigma}_{r}$ and recovered post-convolution $\boldsymbol{\mu}_{r}$ are used for normalization in testing, as demonstrated in \cref{fig4.1}. Besides, we also store task-specific $\boldsymbol{\mu'}_{r}$ like $f_{BN}$'s $\boldsymbol{\mu}_{r}$. Note that Xconv BN just modifies the running moments of BN, so Xconv BN still uses post-convolution $\boldsymbol{\mu}_{BN}$ and $\boldsymbol{\sigma}_{BN}$ in training just as BN does.

\subsection{Hierarchical Fine-tuning}

Xconv BN adjusts the running means to match the shifted representations, i.e., adapts $f_{BN}$ to changed $f_{Conv}$, rather than correcting the intermediate representational shift itself, which remains an issue. 

Firstly, we give an intuition to prevent IRS during fine-tuning. In a common fine-tuning manner, the classification head is randomly initialized and jointly trained with the pre-trained feature extractor. However, since the random head may be far from its optimal value, the loss will be large at the beginning, which also misleads the feature extractor and forces it to change a lot. If we just fine-tune on a single task, it will not cause a problem. But if we fine-tune continually, the massively changed feature extractor may lead to IRS on previous tasks.

To formulate this intuition, we now analyze a simple \emph{overparametrized model} with linear feature extractor following previous work \cite{lpft2,lpft}. Consider a regression task $y = \boldsymbol{v}^\top \boldsymbol{B}\boldsymbol{x}$, in which $\boldsymbol{v}\in\mathbb{R}^k$ is classifier and $\boldsymbol{B}\in\mathbb{R}^{k\times d}$ is feature extractor, and the number of inputs is $n$ which satisfies $1 \leq k < n < d$. We use $\boldsymbol{v}_t^*$ and $\boldsymbol{B}_t^*$ to denote a zero-loss minimum point of task $\mathcal{T}_t$. The MSE loss of task $\mathcal{T}_t$ is denoted as $\mathcal{L}_t(\boldsymbol{B}, \boldsymbol{v})$. For a previous task $\mathcal{T}_{t'}$, since its data could be arbitrary, we consider the worst case in which $\mathcal{L}_{t'}(\boldsymbol{B}, \boldsymbol{v})=\max_{\Vert \boldsymbol{x}\Vert\leq 1}\left( \boldsymbol{v}^\top \boldsymbol{B}\boldsymbol{x} - {\boldsymbol{v}_{t'}^*}^\top \boldsymbol{B}_{t'}^*\boldsymbol{x}\right)^2$. For simplicity, we first analyze the loss of previous tasks in single-head setting, i.e., the model is initialized with $(\boldsymbol{B}_{t-1}^*, \boldsymbol{v}_{t-1}^*)$ and finally updated to $(\boldsymbol{B}_{t}^*, \boldsymbol{v}_{t}^*)$. We here give a proposition which is a deduction of the theorem in \cite{lpft}.

\begin{proposition}
Let $X=\left\{\boldsymbol{x}|\boldsymbol{x}\in\mathcal{T}_{t}\right\}$ be the training data in $\mathcal{T}_{t}$, $\mathcal{S}$ and $\mathcal{R}$ be the orthogonal basis of $SpanSpace(X)$ and $RowSpace(\boldsymbol{B}_{t-1}^*)$ respectively, and $(\boldsymbol{B}_*, \boldsymbol{v}_*)$ be the optimal parameters on both current task $\mathcal{T}_t$ and previous task $\mathcal{T}_{t'}$. If $\sigma_k\left(\mathcal{R}^\top \mathcal{S}^\perp\right)>0$, after fine-tuning on $\mathcal{T}_{t}$, the loss on $\mathcal{T}_{t'}$ is lower bounded as
\begin{align}\sqrt{\mathcal{L}_{t'}\left(\boldsymbol{B}_{t}^*, \boldsymbol{v}_{t}^*\right)}\geq \frac{\sigma_k\left(\mathcal{R}^\top \mathcal{S}^\perp\right)}{\sqrt{k}}\frac{\min(\phi,\phi^2/\left\Vert \boldsymbol{B}_*\boldsymbol{v}_*\right\Vert_2)}{\left(1+\left\Vert \boldsymbol{B}_*\boldsymbol{v}_*\right\Vert_2\right)^2}-\epsilon\end{align}
where $\sigma_k$ denotes the k-th largest singular value, $\phi^2=\left\vert\left( {\boldsymbol{v}_{t-1}^*}^\top \boldsymbol{v}_*\right)^2-\left(\boldsymbol{v}_*^\top \boldsymbol{v}_*\right)^2\right\vert$ is the alignment between $\boldsymbol{v}_{t-1}^*$ and $\boldsymbol{v}_*$, and $\epsilon=\min_{\boldsymbol{U}} \left\Vert\boldsymbol{B}_{t-1}^*-\boldsymbol{U}\boldsymbol{B}_*\right\Vert^2_2$ is the distance between $\boldsymbol{B}_{t-1}^*$ and $\boldsymbol{B}_*$ under a rotation.

\end{proposition}

This proposition can be easily generalized to multi-head setting, which will be introduced in Appendix. This proposition indicates that the loss of previous tasks is lower-bounded by the alignment between the initial and optimal classification head. Moreover, the bound is tighter when the feature extractor is better, such as pre-trained as we use. It indicates that it is important to use a ``good'' initialization for the classification head when using a pre-trained feature extractor. 
\begin{figure}[t]
\centering
\includegraphics[height=0.22\textwidth]{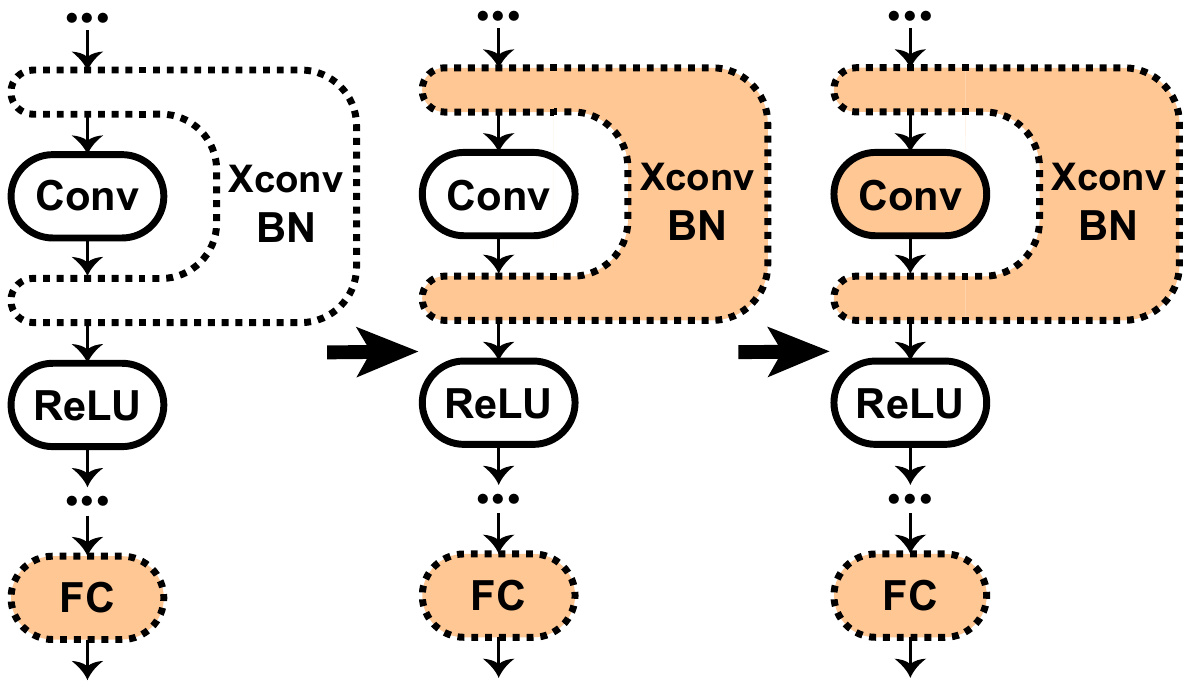}
\caption{Hierarchical fine-tuning. Colored layers are trainable and dashed layers are task-specific.}
\label{fig4.2}
\end{figure}

Along this line of thinking, we propose to fine-tune the classification head first (a.k.a. linear probing) before fine-tuning the entire network, which provides a well-initialized head. To be more specific, when task $\mathcal{T}_{t}$ arrives, we freeze the feature extractor $\boldsymbol{B}_{t-1}^*$ and only fine-tune the head to $\boldsymbol{v}_t^{lp}$. Then the model is initialized with $\left(\boldsymbol{B}_{t-1}^*, \boldsymbol{v}_t^{lp}\right)$ and fine-tune to the optimal point $(\boldsymbol{B}_{t}^{*}, \boldsymbol{v}_{t}^{*})$ on $\mathcal{T}_{t}$. We show that if pre-trained $\boldsymbol{B}_{0}^*$ is ``good enough'', fine-tuning on new task with this proposed strategy will not degrade the performance on previous tasks in multi-head setting.

\begin{proposition}
If every task $\mathcal{T}_{t'}$ with $t'\leq t$ satisfies: (i) while training on this task, $\boldsymbol{v}$ is initialized with $\boldsymbol{v}_{t'}^{lp}$, and (ii) there exists $\boldsymbol{v}_0$ such that $\mathcal{L}_{t'}(\boldsymbol{B}_{0}^*, \boldsymbol{v}_0)=\mathcal{L}_{t'}(\boldsymbol{B}_{t'}^*, \boldsymbol{v}_{t'}^*)$ (i.e., $\boldsymbol{B}_{0}^*$ is good enough), then for all task $\mathcal{T}_{t'}$ with $t'\leq t$, 
\begin{align}\mathcal{L}_{t'}(\boldsymbol{B}_{t}^{*}, \boldsymbol{v}_{t'}^*)=\mathcal{L}_{t'}(\boldsymbol{B}_{t'}^{*}, \boldsymbol{v}_{t'}^*)\end{align}
\end{proposition}

Generalized from the overparametrized model, we can fine-tune ordinary deep networks with a similar strategy. Since Xconv BN stores task-specific moment estimates $\boldsymbol{\mu}$, $\boldsymbol{\sigma}$ and affine parameters $\boldsymbol{\gamma}$, $\boldsymbol{\beta}$ for each task, we can also regard a Xconv BN layer as a ``head'' of a network fragment $f_{BN}\circ f_{Conv}$, and hope that the Xconv BN layers are well-initialized as well, so as to reduce changes of $f_{Conv}$. Consequently, we also fine-tune the Xconv BN layers before fine-tuning the whole network, which finally raises a three-step fine-tuning strategy illustrated in \cref{fig4.2}: when a new task arrives, we first fine-tune  the classification head, and then the head as well as Xconv BNs, and finally all parameters.

\begin{table*}
\centering
\scalebox{0.82}{
\begin{tabular}{p{3cm}<{}p{1.5cm}<{\centering}p{1.5cm}<{\centering}p{1.5cm}<{\centering}p{1.5cm}<{\centering}p{1.5cm}<{\centering}p{1.5cm}<{\centering}p{1.5cm}<{\centering}p{1.5cm}<{\centering}}
\toprule[1.0pt]
\multirow{2}{*}{\textbf{Methods}}  & \multicolumn{2}{c}{CIFAR100 \tiny{$(T=20)$}}&\multicolumn{2}{c}{CUB200 \tiny{$(T=10)$}}&\multicolumn{2}{c}{Caltech101 \tiny{$(T=10)$}}&\multicolumn{2}{c}{Flowers102 \tiny{$(T=10)$}} \\
\specialrule{0em}{1pt}{1pt}

\cline{2-9}

\specialrule{0em}{1pt}{1pt}
&ACC $\uparrow$&FGT $\downarrow$&ACC $\uparrow$&FGT $\downarrow$&ACC $\uparrow$&FGT $\downarrow$&ACC $\uparrow$&FGT $\downarrow$\\

\specialrule{1pt}{1pt}{1pt}

\specialrule{0em}{1pt}{1pt}

Fine-tuning      &  87.43$_{\pm0.78}$ &7.11$_{\pm0.92}$    &71.76$_{\pm0.58}$ &14.76$_{\pm0.66}$&71.89$_{\pm0.86}$ &17.24$_{\pm1.11}$ &78.06$_{\pm1.29}$&10.93$_{\pm1.16}$    \\\specialrule{0em}{1pt}{1pt}
EWC \small{\cite{ewc}}       & 88.04$_{\pm0.59}$  & 6.40$_{\pm0.58}$   &79.30$_{\pm0.43}$ &7.20$_{\pm0.52}$&76.10$_{\pm1.12}$ &12.35$_{\pm1.46}$ &79.94$_{\pm0.68}$&8.56$_{\pm0.56}$   \\\specialrule{0em}{1pt}{1pt}
SI \small{\cite{si}}    & 88.20$_{\pm0.44}$  & 6.42$_{\pm0.46}$&77.94$_{\pm0.53}$&8.42$_{\pm0.51}$&76.26$_{\pm1.32}$& 12.28$_{\pm1.10}$ &79.61$_{\pm0.62}$&8.77$_{\pm0.36}$ \\\specialrule{0em}{1pt}{1pt}
RWalk \small{\cite{rwalk}}   & 87.75$_{\pm0.74}$  &6.82$_{\pm0.78}$ &77.27$_{\pm0.64}$&9.24$_{\pm0.64}$&76.14$_{\pm1.15}$&12.51$_{\pm1.03}$&79.48$_{\pm0.61}$& 9.06$_{\pm0.68}$ \\\specialrule{0em}{1pt}{1pt}
MAS \small{\cite{mas}}  &90.59$_{\pm0.35}$&3.61$_{\pm0.28}$&83.64$_{\pm0.39}$&2.42$_{\pm0.44}$&81.76$_{\pm1.60}$& 6.18$_{\pm0.96}$& 80.90$_{\pm0.82}$  &  6.60$_{\pm0.76}$ \\\specialrule{0em}{1pt}{1pt}

CPR \small{\cite{cpr}}   & 91.17$_{\pm0.21}$ & 2.89$_{\pm0.16}$ &83.64$_{\pm0.36}$&2.49$_{\pm0.36}$&81.85$_{\pm1.47}$&5.68$_{\pm0.86}$& 80.97$_{\pm0.90}$&6.53$_{\pm0.92}$    \\\specialrule{0em}{1pt}{1pt}
AFEC \small{\cite{afec}}   & 90.68$_{\pm0.11}$  &3.55$_{\pm0.04}$ &83.70$_{\pm0.31}$&2.60$_{\pm0.35}$&82.55$_{\pm0.89}$&5.34$_{\pm0.72}$&81.26$_{\pm0.64}$&     6.38$_{\pm0.57}$ \\\specialrule{0em}{1pt}{1pt}

\textbf{ConFiT (Ours)} &\textbf{92.02$_{\pm0.21}$}&\textbf{2.72$_{\pm0.29}$}&\textbf{87.43$_{\pm0.16}$}&\textbf{1.31$_{\pm0.38}$}&\textbf{88.73$_{\pm0.34}$}&\textbf{0.67$_{\pm0.27}$}&\textbf{86.99$_{\pm0.25}$}&\textbf{1.79$_{\pm0.49}$}\\
\specialrule{0em}{1pt}{1pt}

\hline
\specialrule{0em}{1pt}{1pt}
CCLL \small{\cite{ccll}}   &77.59   &0.00 &44.66&0.00&63.18&0.00&64.13&0.00  \\\specialrule{0em}{1pt}{1pt}
LP     & 91.48$_{\pm0.12}$ & 0.00    &86.03$_{\pm0.08}$ &0.00&77.07$_{\pm0.40}$ &0.00 &71.14$_{\pm0.72}$&0.00    \\\specialrule{0em}{1pt}{1pt}
STL& 93.50$_{\pm0.11}$  &0.00 &86.60$_{\pm0.15}$&0.00&86.65$_{\pm0.69}$&0.00&85.96$_{\pm0.50}$& 0.00\\
\bottomrule[1.0pt]

\end{tabular}
}
\caption{Experimental results on benchmark datasets. Methods in the upper part permit update of shared parameters among tasks. For all methods except CCLL, we report results averaged over 5 runs and their 95\% confidence intervals. $\uparrow$ and $\downarrow$ stand for higher is better and lower is better, respectively.}
\label{table1}
\end{table*}
\begin{figure*}[t]
\centering
\includegraphics[height=0.19\textwidth]{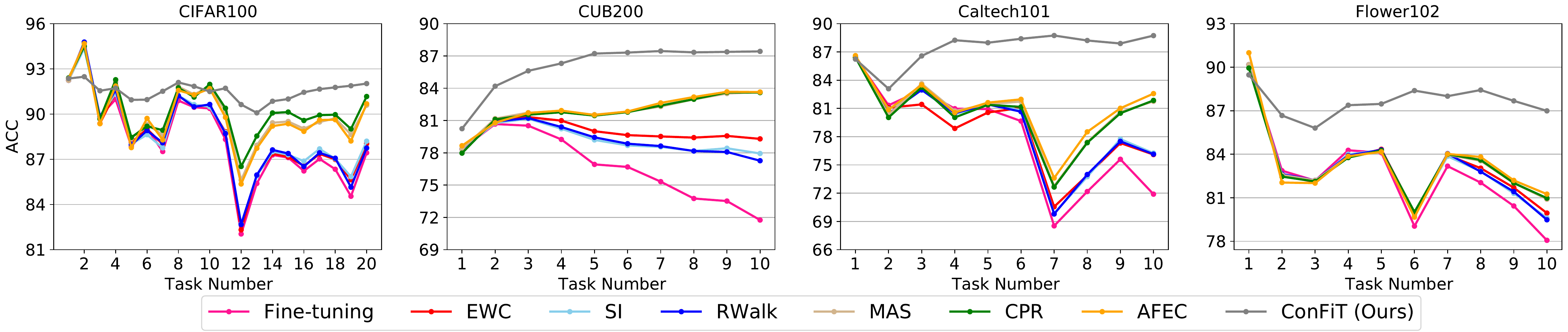}
\caption{Average Accuracy on all already learned tasks after learning each task. We report results averaged over 5 runs.}
\label{fig5.0}
\end{figure*}

\section{Experiments}
\subsection{Setup}
\subsubsection{Datasets}
We evaluate our method on four datasets: CIFAR100 \cite{cifar}, CUB200 \cite{cub}, Caltech101 \cite{caltech}, and Flowers102 \cite{flowers}. CIFAR100 is the most widely used dataset in continual learning, which we also adopt in our experiments. However, CIFAR100 is somewhat easy for a pre-trained model, so we also use three harder datasets: CUB200, which is fine-grained; Caltech101, which is unbalanced; and Flowers102, both fine-grained and unbalanced.

We randomly divide CIFAR100 into 20 tasks where each has 5 classes, and divide CUB200 into 10 tasks where each has 20 classes. For Caltech101 and Flowers102, 100 classes of each dataset are randomly selected and evenly split into 10 tasks. We do not involve Imagenet \cite{imagenet} as a benchmark dataset since it has already been leveraged for pre-training in our experiments. For a fair and clear comparison, the task partitioning of each dataset, as well as its train-test split, is fixed in all runs.

\subsubsection{Baselines}
Since our method does not need episodic memory to store samples and thus is orthogonal to replay-based methods, we compare our method with baselines that also do without episodic memory, including regularization-based methods: EWC \cite{ewc}, SI \cite{si}, RWalk \cite{rwalk}, MAS \cite{mas}, 
CPR \cite{cpr}, and AFEC \cite{afec}; and a parameter-isolation method: CCLL \cite{ccll}. We also report two reference results: single-task learning (STL), where each task is fine-tuned on an individual pre-trained network, and linear probing (LP), where the pre-trained feature extractor is frozen. Note that STL is not a continual learning method.
\subsubsection{Implementation Details}
In all experiments, including all baselines, LP, STL, and ConFiT, we use a ResNet18 pre-trained on ImageNet as initialization. For all regularization-based baselines and our ConFiT, we use a SGD optimizer with a learning rate of 0.01, and train the network with a mini-batch size of \{128, 32, 32, 32\} for \{10, 50, 20, 10\} epochs on each task of \{CIFAR100, CUB200, Caltech101, Flowers102\}, respectively. For CPR and AFEC which are plug-in approaches, we report the performance of MAS-CPR and MAS-AFEC, because MAS performs the best among \{EWC, SI, RWalk, MAS\}. As for our hierarchical fine-tuning, we allocate 20\%, 30\% and 50\% of the total epochs to the three stages, respectively. For parameter-isolation CCLL, we follow the setting in its original paper that trains the network for 150 epochs using SGD with an initial learning rate of 0.01, and multiplies the learning rate by 0.1 at 50, 100 and 125 epochs. All experiments are in multi-head setting, i.e.,  task-incremental scenario. 

%
\subsubsection{Evaluation Metrics}
We measure the performance with two metrics, as in previous works \cite{rwalk,cpr}: Average Accuracy (ACC) and Average Forgetting (FGT). We denote the accuracy on task $\mathcal{T}_{i}$ after training the model on task $\mathcal{T}_{j}$ as $a_{ij}$, and task $\mathcal{T}_{T}$ is the final task. Then ACC and FGT can be calculated as follows:
\begin{align}ACC&=\frac{1}{T}\sum_{i=1}^Ta_{iT}\\
FGT&=\frac{1}{T-1}\sum_{i=1}^{T-1}\max_{i\leq j\leq T}(a_{ij}-a_{iT})\end{align}

\subsection{Results \& Discussion}
\subsubsection{Comparisons with Baselines}
The main results are shown in\cref{table1}. MAS performs better than EWC, SI, and RWalk on all datasets, which is reasonable since MAS will not underestimate the importance of parameters for pre-trained knowledge, as \cite{mas} indicates. Plug-in methods CPR and AFEC do not provide much improvement on MAS in this setting. CCLL performs poorly since the added calibration modules significantly change the structure of the pre-trained network and thus disrupt the pre-trained knowledge.

On all the four datasets, ConFiT has higher ACC than all continual learning baselines, and lower FGT than all regularization-based methods. Furthermore, ConFiT outperforms STL on CUB200, Caltech101, and Flowers102. Since STL uses an individual network for each task, it can achieve zero forgetting, but will not benefit from task-level knowledge transfer. It indicates that ConFiT also has the ability to share knowledge among tasks. On CUB200 and CIFAR100, LP even outperforms many baselines and has good results close to STL. This reflects the great ability of the pre-trained model --- it has accumulated almost enough knowledge during pre-training. Whereas ConFiT still performs better than LP, which indicates the necessity to obtain more task-specific knowledge via fine-tuning.

\begin{figure}[t]
\centering
\includegraphics[height=0.17\textwidth]{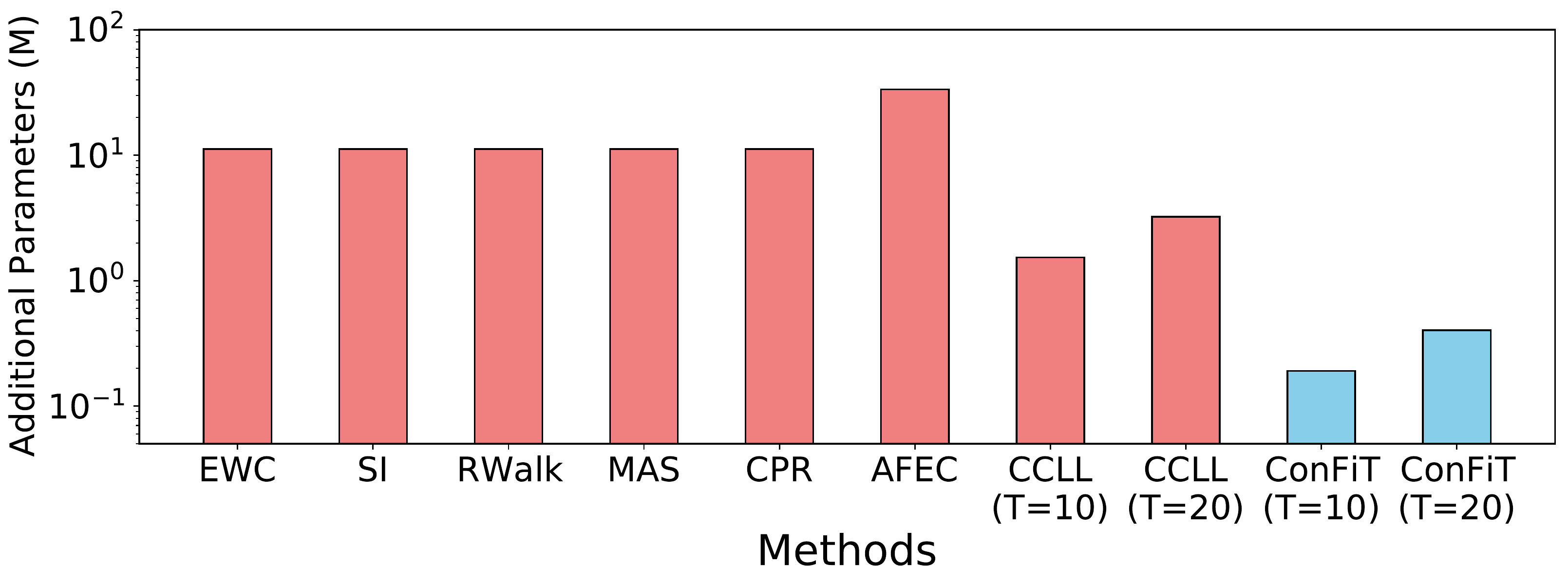}
\caption{Additional parameters to be stored for each method. ConFiT uses orders of magnitude fewer additional parameters than other baselines.}
\label{fig5.1}
\end{figure}

In \cref{fig5.1} we illustrate that ConFiT uses orders of magnitude fewer additional parameters than other baselines. Methods like EWC need to store regularization weights for each parameter (11.2M for ResNet18). In addition to these, AFEC needs to store an extra network and its regularization weights. CCLL stores parameters of calibration modules (0.17M/task). ConFiT, on the other hand,  only stores Xconv BN parameters for each task (0.02M/task). It is worth noting that 0.02M/task is an extremely tiny requirement for continual learning, since the episodic memory will take up about 0.15M/task even if only a single 224$\times$224$\times$3 image/task is stored for replay-based methods.
\begin{table}
\centering
\scalebox{0.86}{
\begin{tabular}{p{2.3cm}<{\centering}p{1.6cm}<{\centering}p{1.9cm}<{\centering}p{1.9cm}<{\centering}}

\toprule[1.0pt]
 \multicolumn{2}{c}{Ablation Cases}&Flowers102&Caltech101\\

\specialrule{1pt}{1pt}{1pt}
\specialrule{0em}{1pt}{1pt}

Fine-tuning&BN&78.06$\pm$1.29&71.89$\pm$0.86\\
\specialrule{0em}{1pt}{1pt}

Hierarchical FT&BN&83.39$\pm$0.43&85.12$\pm$0.34\\
\specialrule{0em}{1pt}{1pt}

Fine-tuning&Xconv BN&81.26$\pm$0.72&82.09$\pm$0.59\\
\specialrule{0em}{1pt}{1pt}

Hierarchical FT&Xconv BN&\textbf{86.99$\pm$0.25}&\textbf{88.73$\pm$0.34}\\

\bottomrule[1.0pt]

\end{tabular}
}
\caption{Ablation study for ConFiT. We report ACC on Flowers102 and Caltech101.}

\label{table2}
\end{table}
\subsubsection{Ablation Study}
We provide an ablation study on Flowers102 and Caltech101. As shown in \cref{table2}, both hierarchical fine-tuning and Xconv BN can boost the average accuracy significantly. ConFiT combines both of them and achieves the best performance.

\begin{figure}[t]
\centering
\includegraphics[height=0.45\textwidth]{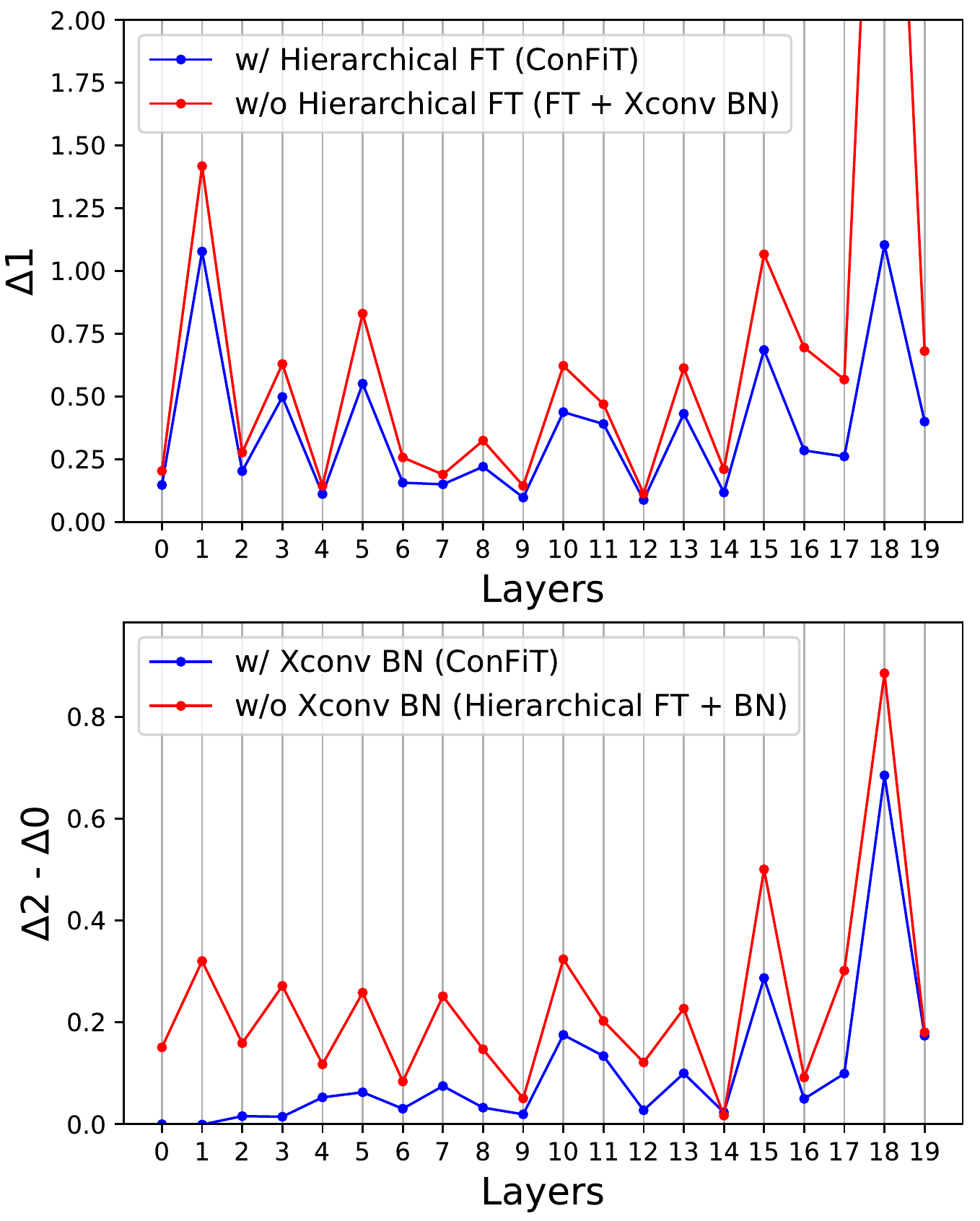}
\caption{$\Delta_1$ and $\Delta_2-\Delta_0$ of the 20 Conv layers in ResNet18 (three of which are in shortcuts) on Flowers102. Layer 0 is closest to input. Lower $\Delta_1$ and $\Delta_2-\Delta_0$ are better.}
\label{fig5.4}
\end{figure}

Moreover, to verify our motivation --- ConFiT alleviates IRS and corrects the BN running estimates, we record these statistics for output of each Conv layer: 
\begin{itemize}
    \item $\boldsymbol{\mu}_{te}^i$:  the true mean on test set of $\mathcal{T}_1$ after training on task $\mathcal{T}_i$;
    \item $\boldsymbol{\mu}_{r}^i$:  the post-convolution running mean of BN/XConv BN of $\mathcal{T}_1$ after training on $\mathcal{T}_i$. 
\end{itemize}
Note that for BN the running mean is shared across tasks, whereas for Xconv BN it is specific for $\mathcal{T}_1$. All the means are calculated along the axes of $BHW$ as BN does. Based on them, we calculate the following values on Flowers102: 
\begin{itemize}
    \item $\Delta_1 = \Vert\boldsymbol{\mu}_{te}^1-\boldsymbol{\mu}_{te}^{10}\Vert_2$: the shift of the mean of $\mathcal{T}_1$'s intermediate representations;
    \item $\Delta_2 =\Vert\boldsymbol{\mu}_{r}^{10}-\boldsymbol{\mu}_{te}^{10}\Vert_2$:  the gap between the  running mean and true mean of $\mathcal{T}_1$ when the training is over. 
\end{itemize}
Note that the running estimates are not absolutely precise even with Xconv BN, since Xconv BN inherits the intrinsic inconsistency between running and true moments of BN. For a clear comparison, we also calculate this intrinsic inconsistency as
\begin{itemize}
\item $\Delta_0 =\Vert\boldsymbol{\mu}_{r}^1-\boldsymbol{\mu}_{te}^1\Vert_2$:  the gap between the running mean and true mean of $\mathcal{T}_1$ when the network has only been trained on $\mathcal{T}_1$, which is a lower bound of $\Delta_2$. 
\end{itemize}
We conduct experiments on ablation cases where hierarchical fine-tuning and Xconv BN are not used, and report the $\Delta_1$ and $\Delta_2-\Delta_0$ of each layer.

\cref{fig5.4} shows that, for the majority of layers, IRS has been alleviated by hierarchical fine-tuning (lower $\Delta_1$), and error of estimation has been reduced by Xconv BN (lower $\Delta_2-\Delta_0$). Moreover, the IRS of the final Conv layer's output also decreases via hierarchical fine-tuning, which means the representational shift of features is also reduced. We also find that Xconv BN is more effective on the shallow layers (close to input), where the estimation error of Xconv BN is almost 0. This is because the estimation error may accumulate layer by layer, which makes the inputs' means of deep layers not be fixed, potentially violating the assumption in \cref{s3.2}. 

\subsubsection{Is Running Variance Correction Necessary?}
\label{s4.2}
Xconv BN maintains pre-convolution running means instead of post-convolution, but it still retains the post-convolution running variances like BN. A straightforward reason for using post-convolution running variances is that the variance is a nonlinear secondary moment, so it is difficult to find a feasible mapping between pre- and post-convolution variances under a linear Conv layer.

\begin{figure}[t]
\centering
\includegraphics[height=0.14\textwidth]{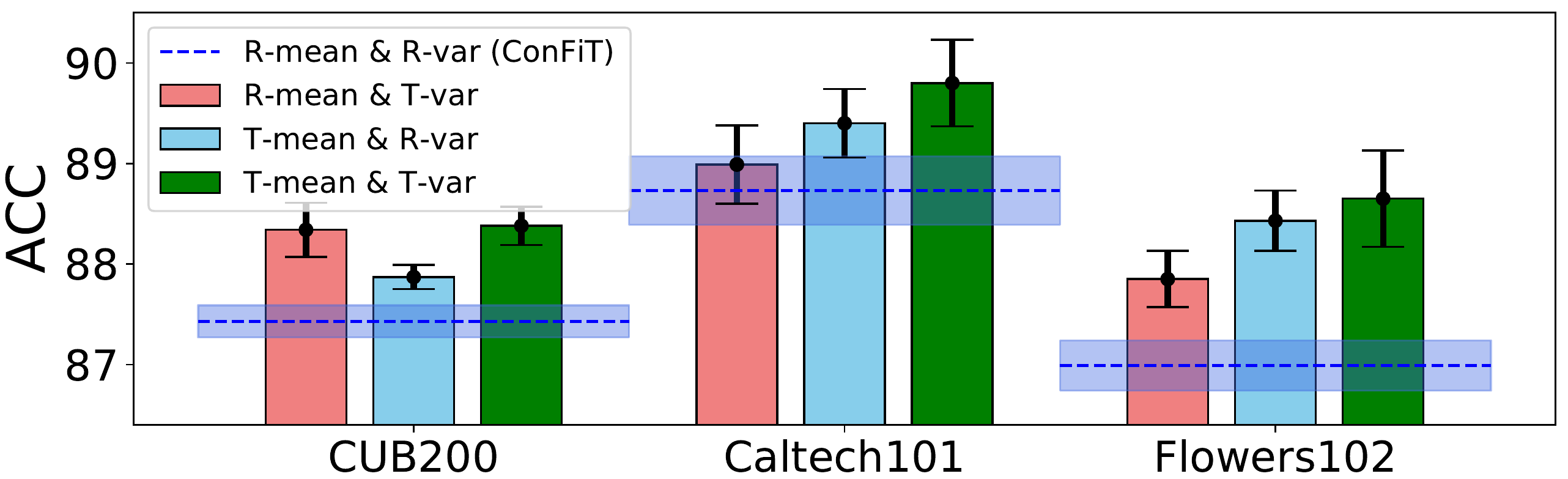}
\caption{Results when using transductive moment estimates. ``R-'' and ``T-'' stand for ``running'' and ``transductive'', respectively. The R-mean has been corrected by Xconv BN.}
\label{fig5.2}
\end{figure}

To figure out whether running variance correction is necessary, we explore the best case of correction: using true moments to normalize. We refer to this ideal setting as ``transductive'' since the model has access to the whole test set to determine parameters. As shown in \cref{fig5.2}, if the mean and variance are both transductive, the normalization is accurate, so the model outperforms ConFiT as expected, whose mean and variance are both running. If only the mean or only the variance is transductive, the performance is also improved. However, the improvement brought about by transductive variance is somewhat limited (less than 1\%) compared with the large improvement that ConFiT has already provided, not to mention that the transductive results are just upper bounds in the ideal cases. Therefore, we argue that running variance correction is insignificant for Xconv BN. 

\subsubsection{Performance on Deeper Networks}

Since our experiments are conducted on ResNet18, we here deploy ConFiT on  two other networks: ResNet34  and ResNet50 \cite{resnet}, which are much deeper than ResNet18. These networks also use BN after Conv layers, so our ConFiT can be directly applied. Since deeper networks converge more slowly, we train each network for 50 epochs on Caltech101 and Flowers102, more than that for ResNet18. 

\begin{figure}[t]
\centering
\includegraphics[height=0.16\textwidth]{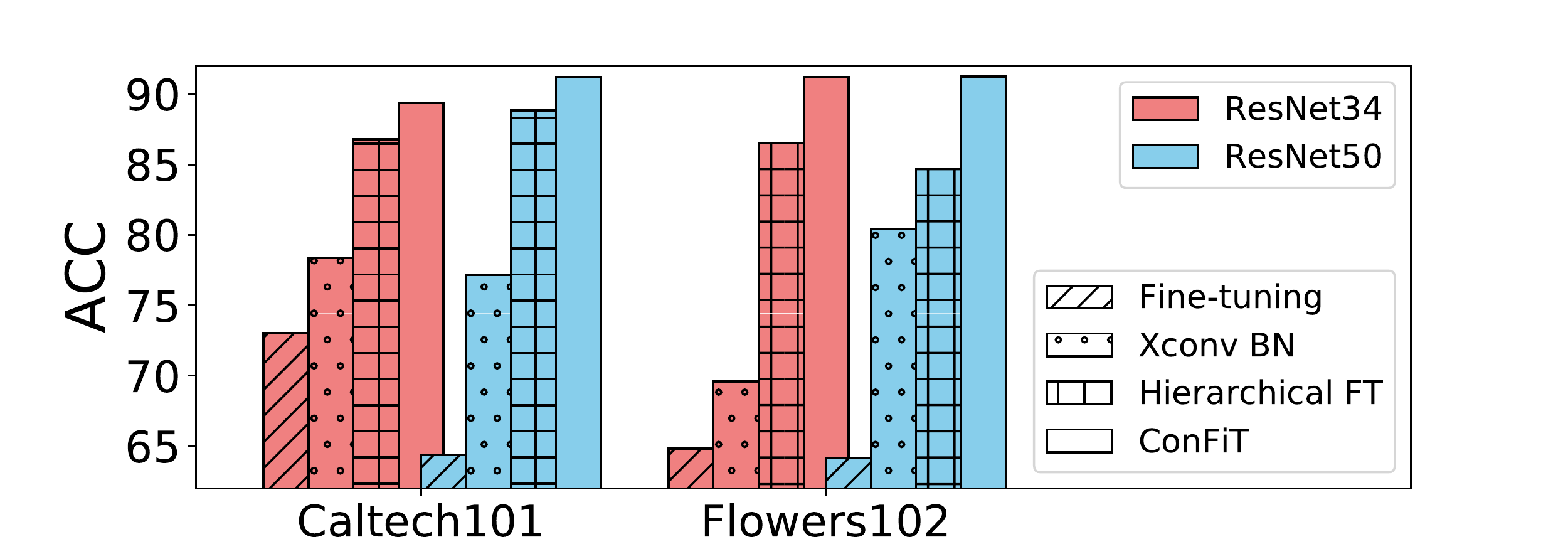}
\caption{ACC on ResNet34 and ResNet50.}
\label{app_fig2}
\end{figure}

As shown in \cref{app_fig2}, when directly fine-tuning the network, both ResNet34 and ResNet50 perform poorly, and the deeper ResNet50 even suffers from more forgetting. ConFiT boosts the performance of both ResNet34 and ResNet50, in which the ACC of ResNet50 with ConFiT is higher than that of ResNet34. This illustrates that ConFiT has great scalability and can be applied to deeper networks according to actual demand.


\section{Limitations}
Firstly, the proposed hierarchical fine-tuning is based on a strong assumption that the feature extractor has already been good enough before continual fine-tuning. This assumption makes it hard to generalize this method to a randomly initialized network, or to cases when the tasks are too complicated for the pre-trained model to handle well.

Secondly, Xconv BN needs to store a small number of task-specific parameters, so the task identifiers are necessary in testing for selecting the parameters of Xconv BN. Consequently, Xconv BN cannot be used for class-incremental learning and task-agnostic scenarios directly. Xconv BN can also not be applied to models without BN, such as Vision Transformers \cite{vit}. 

\section{Conclusion}
In this paper, we focus on fine-tuning continually on pre-trained models. We reveal a crucial cause of catastrophic forgetting --- the IRS distorts BN. From this perspective, we propose ConFiT, which corrects running means via Xconv BN and alleviates IRS via hierarchical fine-tuning. These two components are proven to be effective both empirically and theoretically. Our ConFiT achieves superior performance, surpassing current SOTA regularization-based and parameter-isolation methods on several datasets.

\small{
\bibliographystyle{ieee_fullname}
\bibliography{ijcai22}
}
\newpage
\section*{Appendix}

\setcounter{section}{0}

\renewcommand\thesection{\Alph{section}}
\section{Proofs}
\subsection{Proof of Proposition 1}
\emph{Proof}. 
Without loss of generality, we suppose the Conv layer does not have a bias parameter. We denote its kernel size as $K$, and its weight parameter as $\boldsymbol{W}\in\mathbb{R}^{C\times C'\times K\times K}$. When padding size is $K-1$, the input $\boldsymbol{a}\in\mathbb{R}^{B\times C\times H\times W}$ is zero-padded to $\boldsymbol{a}'\in\mathbb{R}^{B\times C\times (H+2K-2)\times (W+2K-2)}$. The output of this Conv layer is $f_{Conv}(\boldsymbol{a})\in\mathbb{R}^{B\times C'\times H'\times W'}$. We have:
\begin{align*}
&AvgPool(f_{Conv}(\boldsymbol{a}))\\
=&\frac{1}{BH'W'}\sum_{b=1}^B\sum_{h=1}^{H'}\sum_{w=1}^{W'}f_{Conv}(\boldsymbol{a})_{b,:,h,w}\\
=&\frac{1}{BH'W'}\sum_{b=1}^B\sum_{h=1}^{H+K-1}\sum_{w=1}^{W+K-1}\\
&\quad\quad\quad\quad\quad\quad\sum_{c=1}^{C}\sum_{i=1}^{K}\sum_{j=1}^{K}\boldsymbol{W}_{c,:,i,j}\boldsymbol{a}'_{b,c,h+i-1,w+j-1}\\
=&\frac{1}{BH'W'}\sum_{c=1}^{C}\sum_{i=1}^{K}\sum_{j=1}^{K}\Bigg(\boldsymbol{W}_{c,:,i,j}\\
&\quad\quad\quad\quad\quad\quad\sum_{b=1}^B\sum_{h=1}^{H+K-1}\sum_{w=1}^{W+K-1}\boldsymbol{a}'_{b,c,h+i-1,w+j-1}\Bigg)\\
=&\frac{1}{BH'W'}\sum_{c=1}^{C}\sum_{i=1}^{K}\sum_{j=1}^{K}\left(\boldsymbol{W}_{c,:,i,j}\sum_{b=1}^B\sum_{h=1}^{H}\sum_{w=1}^{W}\boldsymbol{a}_{b,c,h,w}\right)\\
=&\frac{1}{BH'W'}\sum_{c=1}^{C}\sum_{i=1}^{K}\sum_{j=1}^{K}\Bigg(\boldsymbol{W}_{c,:,i,j}\\
&\quad\quad\quad\quad\quad\quad\sum_{b=1}^B\sum_{h=1}^{H}\sum_{w=1}^{W}AvgPool_{DP}(\boldsymbol{a})_{b,c,h,w}\Bigg)\\
=&\frac{1}{BH'W'}\sum_{c=1}^{C}\sum_{i=1}^{K}\sum_{j=1}^{K}\Bigg(\boldsymbol{W}_{c,:,i,j}\\
&\quad\sum_{b=1}^B\sum_{h=1}^{H+K-1}\sum_{w=1}^{W+K-1}AvgPool_{DP}(\boldsymbol{a})'_{b,c,h+i-1,w+j-1}\Bigg)\\
=&\frac{1}{BH'W'}\sum_{b=1}^B\sum_{h=1}^{H'}\sum_{w=1}^{W'}f_{Conv}(AvgPool_{DP}(\boldsymbol{a}))_{b,:,h,w}\\
=&AvgPool(f_{Conv}(AvgPool_{DP}(\boldsymbol{a})))
\end{align*}

\begin{figure}[h]
\centering
\includegraphics[height=0.20\textwidth]{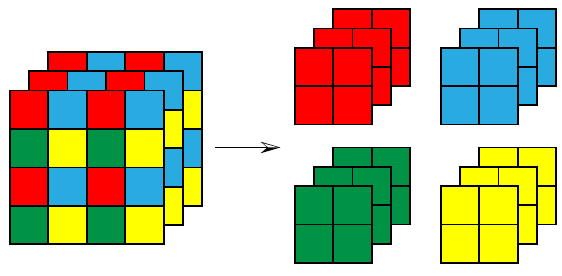}
\caption{}
\label{app_fig1}
\end{figure}
If $stride\neq 1$, we suppose $stride=2$ as an instance. We can transform the convolution with $stride=2$ into 4 convolutions with $stride=1$, by breaking down both $\boldsymbol{a}$ and $\boldsymbol{W}$ as in \cref{app_fig1}. Then the proposition still holds for each of the 4 convolutions. Meanwhile, we need to store 4 pre-convolution means with respect to the 4 parts of $\boldsymbol{a}$. Similarly, if $stride=m$, we just need to transform the convolution into $m^2$ convolutions with $stride=1$,  and store $m^2$ pre-convolution means.

\begin{table*}[th]
\centering
\scalebox{1}{
\begin{tabular}{p{7cm}<{\centering}p{2cm}<{\centering}p{2cm}<{\centering}p{2cm}<{\centering}p{2cm}<{\centering}}
\toprule[1.0pt]
\textbf{Dataset}&CIFAR100&CUB200&Caltech101&Flowers102\\
\specialrule{1pt}{1pt}{1pt}
\specialrule{0em}{1pt}{1pt}
\#classes&100&200&101&102\\
\specialrule{0em}{1pt}{1pt}
\#tasks&20&10&10&10\\
\specialrule{0em}{1pt}{1pt}
\#classes in each task&5&20&10&10\\
\specialrule{0em}{1pt}{1pt}
max  \#images of a class in training set&500&30&640&206\\
\specialrule{0em}{1pt}{1pt}
min  \#images of a class in training set&500&29&25&32\\
\specialrule{0em}{1pt}{1pt}
\#images in training set&50000&5994&6941&6551\\
\specialrule{0em}{1pt}{1pt}
\#images in test set&10000&5794&1736&1638\\
\bottomrule[1.0pt]

\end{tabular}
}
\caption{Main statistics of datasets.}
\label{app_table1}
\end{table*}

As for the cases where $padding\neq K-1$, we hypothesize the pixels at the edges of images are less informative background, which can be ignored securely. So the proposition still holds with negligible error.
\subsection{Proof of Proposition 2}
\emph{Proof}. 
This proposition is an immediate deduction of Theorem 3.2 in \cite{lpft} \footnote{We notice that the theorem IDs are different between the openreview camera-ready version and arxiv version. We are in line with the former.} by simply regarding inputs of task $\mathcal{T}_{t}$ as in-distribution data, and inputs of task $\mathcal{T}_{t'}$ as out-of-distribution data. A slight difference is that we assume the ``out-of-distribution data'' are in the deterministic worst case, so we do not have a second moment term $\sqrt{\sigma_{min}(\Sigma)}$ which is derived from the stochasticity of the out-of-distribution data in Lemma A.7 of \cite{lpft}.

As for multi-head setting, we have 
\begin{align*}
&\sqrt{\mathcal{L}_{t'}(\boldsymbol{B}_{t}^*, \boldsymbol{v}_{t'}^*)}=\sqrt{\max_{\Vert \boldsymbol{x}\Vert\leq 1}\left( \boldsymbol{v}_{t'}^{*\top} \boldsymbol{B}_{t}^*\boldsymbol{x} - \boldsymbol{v}_{t'}^{*\top} \boldsymbol{B}_{t'}^*\boldsymbol{x}\right)^2}\\
&=\left\Vert\boldsymbol{v}_{t'}^{*\top} \boldsymbol{B}_{t}^*-\boldsymbol{v}_{t'}^{*\top} \boldsymbol{B}_{t'}^*\right\Vert_2\\
&\geq\left\Vert\boldsymbol{v}_{t}^{*\top} \boldsymbol{B}_{t}^*-\boldsymbol{v}_{t'}^{*\top} \boldsymbol{B}_{t'}^*\right\Vert_2-\left\Vert\boldsymbol{v}_{t}^{*\top} \boldsymbol{B}_{t}^*-\boldsymbol{v}_{t'}^{*\top} \boldsymbol{B}_{t}^*\right\Vert_2\\
&=\sqrt{\mathcal{L}_{t'}(\boldsymbol{B}_{t}^*, \boldsymbol{v}_{t}^*)}-\epsilon_{mh}
\end{align*}
in which $\epsilon_{mh}=\left\Vert\boldsymbol{v}_{t}^{*\top} \boldsymbol{B}_{t}^*-\boldsymbol{v}_{t'}^{*\top} \boldsymbol{B}_{t}^*\right\Vert_2$ is the divergence between the two sub-networks corresponding to the two heads. We can regard $\epsilon_{mh}$ as a relaxation provided by multi-head setting.

\subsection{Proof of Proposition 3}
\emph{Proof}. 
For all task $\mathcal{T}_{t'}$ with  $t'\leq t$, we firstly assume that $\boldsymbol{B}_{t'-1}^*=\boldsymbol{B}_{0}^*$. Then there exists $\boldsymbol{v}_0$ such that $\mathcal{L}_{t'}(\boldsymbol{B}_{t'-1}^*, \boldsymbol{v}_0)=\mathcal{L}_{t'}(\boldsymbol{B}_{t'}^*, \boldsymbol{v}_{t'}^*)$. By the proof of Proposition A.21 in \cite{lpft} we have $\boldsymbol{v}_{t'}^{lp}=\boldsymbol{v}_0$. Since overparametrized model can achieve zero loss, we have $\mathcal{L}_{t'}\left(\boldsymbol{B}_{t'-1}^*, \boldsymbol{v}_{t'}^{lp}\right)=\mathcal{L}_{t'}(\boldsymbol{B}_{0}^*, \boldsymbol{v}_0)=0$. So the gradient is also zero, and thus the feature extractor does not change at all, which means $\boldsymbol{B}_{t'}^*=\boldsymbol{B}_{0}^*$ as well. 

When $t' = 1$, the assumption $\boldsymbol{B}_{t'-1}^*=\boldsymbol{B}_{0}^*$ is naturally satisfied. Then inductively, for all task $\mathcal{T}_{t'}$ with $t'\leq t$, we have $\boldsymbol{B}_{t'}^*=\boldsymbol{B}_{0}^*$, which leads to $\mathcal{L}_{t'}(\boldsymbol{B}_{t}^{*}, \boldsymbol{v}_{t'}^*)=\mathcal{L}_{t'}(\boldsymbol{B}_{t'}^{*}, \boldsymbol{v}_{t'}^*)$ as we desire.

\begin{table*}
\centering
\scalebox{0.9}{
\begin{tabular}{p{1cm}<{\centering}p{2cm}<{\centering}p{7cm}<{\centering}p{1.5cm}<{\centering}p{1.5cm}<{\centering}p{1.5cm}<{\centering}p{1.5cm}<{\centering}}
\toprule[1.0pt]
\textbf{Method}&Hyperparameter&Search Range  & CIFAR100&CUB200&Caltech101&Flowers102  \\
\specialrule{0em}{1pt}{1pt}

\specialrule{1pt}{1pt}{1pt}

\specialrule{0em}{1pt}{1pt}

EWC      & $\lambda$  & \{3e2, 1e3, 3e3, 1e4, 3e4, 1e5, 3e5\}   &1e3 &3e3&3e4 &1e5  \\\specialrule{0em}{1pt}{1pt}
SI    & $c$ & \{1, 2, 5, 10, 20, 50\}   &2 &5&10 &20 \\\specialrule{0em}{1pt}{1pt}
RWalk    & $\lambda$ & \{5, 10, 20, 50, 100, 200\}    &20 &20&50 &100   \\\specialrule{0em}{1pt}{1pt}
MAS &$\lambda$& \{1, 3, 10, 30, 100, 300\}   &10 &30&30 &30  \\\specialrule{0em}{1pt}{1pt}

CPR   & $\beta$ & \{0.005, 0.01, 0.02, 0.05, 0.1, 0.2, 0.5\}   &0.2 &0.02&0.01 &0.1    \\\specialrule{0em}{1pt}{1pt}
AFEC  & $\lambda_e$ & \{0.001, 0.01, 0.1, 1, 10, 100, 1000, 10000\}   &0.01 &100&1000 &10 \\\specialrule{0em}{1pt}{1pt}

\bottomrule[1.0pt]

\end{tabular}
}
\caption{Results of hyperparameter search.}
\label{app_table2}
\end{table*}

\section{Experiment Details}
\subsection{Dataset Split}
For CIFAR100 and CUB200, we use the official train/test split. For Caltech101 and Flowers102, we randomly choose 80\% images as training set, and the others as test set. During hyperparameter search, we further randomly choose 10\% images in training set for validation. The details of dataset statistics are in \cref{app_table1}.
\subsection{Implementation}
All experiments are implemented using \emph{Pytorch} 1.10 with CUDA 10.2 on NVIDIA 1080Ti GPU. The datasets and dataloaders are built via \emph{Avalanche} \cite{Avalanche}. We use ResNet18 implemented by \emph{timm} \cite{timm}, and its pre-trained parameters are downloaded from \emph{torchvision}. The baselines are reproduced on the basis of the official codebases of \cite{cpr,afec,ccll}, in which we only modify the models and dataset interfaces. 
\subsection{Data Pre-processing}
For each image, we pre-process it as follows (Pytorch style):
\begin{verbatim}
Compose(
    Resize(size=256),
    CenterCrop(size=(224, 224)),
    ToTensor(),
    Normalize(
        mean=[0.4850, 0.4560, 0.4060], 
        std=[0.2290, 0.2240, 0.2250]
    )
)
\end{verbatim}
There is no extra data augmentation applied.

\subsection{Hyperparameter Search}
For each baseline, we extensively search the best hyperparameter, which is shown in \cref{app_table2}. We train the model for a fixed number of epochs for each method, which depends on the complexity of the dataset. We randomly choose 10\% images in training set for validation, exclude them from training, and use them to determine the best hyperparameters. After the hyperparameter search, we retrain the model with the whole training set and report the results on test set. Note that our ConFiT does not need hyperparameter search.

\section{Difference between IRS and \emph{Internal Covariate Shift}}
We notice that there is another concept called \emph{Internal Covariate Shift} (ICS) about BN. To avoid confusion, we here clarify the difference between IRS and ICS. 

Firstly, we would emphasize that ICS and IRS are totally different concepts. ICS describes the instability of intermediate representations' distribution during the training stage, i.e., the inputs of intermediate layers are unstable \emph{in training}, which impedes the  learning of networks. One of the motivations behind BN is to address ICS, so as to make the network easier to train.

Whereas IRS is defined in the scenario of continual learning, which means the intermediate representations of \emph{data of previous tasks} were shifted, because the network fitted the data of the newest task. The shifted representation will disrupt the function of BN \emph{in testing}, since the running moments will  no longer be representative of the true moments of intermediate representations.

Overall, BN can solve ICS, but will suffer from IRS in continual learning, which is what we have attempted to address.

\end{document}


\maketitle
\section{Proofs}
\subsection{Proof of Theorem 1}

\begin{theorem} Suppose $Conv(\cdot)$ denotes a 2D-Conv layer with $stride=1$ and $padding=kernel size-1$, and its output has a shape of  $B\times C'\times H'\times W'$. Then:
\begin{align*}
\frac{1}{BH'W'}\sum_{b=1}^B\sum_{h=1}^{H'}\sum_{w=1}^{W'}&Conv(\boldsymbol{a})_{b,:,h,w}=\\
\frac{1}{BH'W'}\sum_{b=1}^B\sum_{h=1}^{H'}\sum_{w=1}^{W'}&Conv(AvgPool_{DP}(\boldsymbol{a}))_{b,:,h,w}
\end{align*}
\end{theorem}
\begin{proof}
Without loss of generality, we suppose the Conv layer does not have bias parameter. We denote its kernel size as $K$, and its weight parameter as $\boldsymbol{W}\in\mathbb{R}^{C\times C'\times K\times K}$. When padding size is $K-1$, the input $a$ is zero-padded to $\boldsymbol{a}'\in\mathbb{R}^{B\times C'\times (H'+2K-2)\times (W'+2K-2)}$. We have:
\begin{align*}
&\frac{1}{BH'W'}\sum_{b=1}^B\sum_{h=1}^{H'}\sum_{w=1}^{W'}Conv(\boldsymbol{a})_{b,:,h,w}\\
=&\frac{1}{BH'W'}\sum_{b=1}^B\sum_{h=1}^{H}\sum_{w=1}^{W}\sum_{c=1}^{C}\sum_{i=1}^{K}\sum_{j=1}^{K}\boldsymbol{W}_{c,:,i,j}\boldsymbol{a}'_{b,c,h+i-1,w+j-1}\\
=&\frac{1}{BH'W'}\sum_{c=1}^{C}\sum_{i=1}^{K}\sum_{j=1}^{K}(\boldsymbol{W}_{c,:,i,j}\\
&\quad\quad\quad\quad\quad\quad\sum_{b=1}^B\sum_{h=1}^{H}\sum_{w=1}^{W}\boldsymbol{a}'_{b,c,h+i-1,w+j-1})\\
=&\frac{1}{BH'W'}\sum_{c=1}^{C}(\sum_{i=1}^{K}\sum_{j=1}^{K}\boldsymbol{W}_{c,:,i,j}\sum_{b=1}^B\sum_{h=1}^{H}\sum_{w=1}^{W}\boldsymbol{a}_{b,c,h,w})\\
=&\frac{1}{BH'W'}\sum_{c=1}^{C}(\sum_{i=1}^{K}\sum_{j=1}^{K}\boldsymbol{W}_{c,:,i,j}\\
&\quad\quad\quad\quad\quad\quad\sum_{b=1}^B\sum_{h=1}^{H}\sum_{w=1}^{W}AvgPool_{DP}(\boldsymbol{a})_{b,c,h,w})\\
\end{align*}
\begin{align*}
=&\frac{1}{BH'W'}\sum_{c=1}^{C}\sum_{i=1}^{K}\sum_{j=1}^{K}(\boldsymbol{W}_{c,:,i,j}\\
&\quad\quad\quad\quad\quad\sum_{b=1}^B\sum_{h=1}^{H}\sum_{w=1}^{W}AvgPool_{DP}(\boldsymbol{a})'_{b,c,h+i-1,w+j-1})\\
=&\frac{1}{BH'W'}\sum_{b=1}^B\sum_{h=1}^{H'}\sum_{w=1}^{W'}Conv(AvgPool_{DP}(\boldsymbol{a}))_{b,:,h,w}
\end{align*}
\end{proof}
\begin{figure}[t]
\centering
\includegraphics[height=0.20\textwidth]{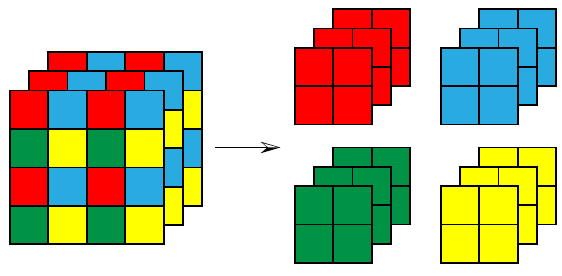}
\caption{}
\label{app_fig1}
\end{figure}
If $stride\neq 1$, we suppose $stride=2$ as an instance. We can transform the convolution with $stride=2$ into 4 convolutions with $stride=1$, by breaking down both $\boldsymbol{a}$ and $\boldsymbol{W}$ as Figure \ref{app_fig1}. Then the theorem still holds for each of the 4 convolutions. Meanwhile, we need to store 4 pre-convolution means with respect to the 4 part of $\boldsymbol{a}$. Similarly, if $stride=m$, we just need to transform the convolution into $m^2$ convolutions with $stride=1$,  and store $m^2$ pre-convolution means.

\begin{table*}[th]
\centering
\scalebox{1}{
\begin{tabular}{p{7cm}<{\centering}p{2cm}<{\centering}p{2cm}<{\centering}p{2cm}<{\centering}p{2cm}<{\centering}}
\toprule[1.0pt]
\textbf{Dataset}&CIFAR100&CUB200&Caltech101&Flowers102\\
\specialrule{1pt}{1pt}{1pt}
\specialrule{0em}{1pt}{1pt}
\#classes&100&200&101&102\\
\specialrule{0em}{1pt}{1pt}
\#tasks&20&10&10&10\\
\specialrule{0em}{1pt}{1pt}
\#classes in each task&5&20&10&10\\
\specialrule{0em}{1pt}{1pt}
max  \#images of a class in training set&500&30&640&206\\
\specialrule{0em}{1pt}{1pt}
min  \#images of a class in training set&500&29&25&32\\
\specialrule{0em}{1pt}{1pt}
\#images in training set&50000&5994&6941&6551\\
\specialrule{0em}{1pt}{1pt}
\#images in test set&10000&5794&1736&1638\\
\bottomrule[1.0pt]

\end{tabular}
}
\caption{Main statistics of datasets.}
\label{app_table1}
\end{table*}

As for the cases where $padding\neq K-1$, we hypothesize the pixels at the edges of images are less informative background, which can be ignored securely, and thus the theorem holds with negligible error.
\subsection{Proof of Theorem 2}
\begin{theorem}
Let $X=\{\boldsymbol{x}|\boldsymbol{x}\in\mathcal{T}_{t}\}$ be the training data, $\mathcal{S}=SpanSpace(X)$ and $\mathcal{R}=RowSpace(\boldsymbol{B}_{t-1}^*)$ be the orthogonal basis of corresponding spaces, and $(\boldsymbol{B}_*, \boldsymbol{v}_*)$ be the optimal parameters on both current task $t$ and previous task $t'$. If $\sigma_k(\mathcal{R}^\top \mathcal{S}^\perp)>0$. After fine-tuning on task $t$, the loss on task $t'$ is lower bounded:
$$\sqrt{\mathcal{L}_{t'}(\boldsymbol{B}_{t}^*, \boldsymbol{v}_{t}^*)}\geq \frac{\sigma_k(\mathcal{R}^\top \mathcal{S}^\perp)}{\sqrt{k}}\frac{min(\phi,\phi^2/\Vert \boldsymbol{B}_*\boldsymbol{v}_*\Vert_2)}{(1+\Vert \boldsymbol{B}_*\boldsymbol{v}_*\Vert_2)^2}-\epsilon$$
where $\sigma_k$ denotes the k-th largest singular value, $\phi^2=\vert( {\boldsymbol{v}_{t-1}^*}^\top \boldsymbol{v}_*)^2-(\boldsymbol{v}_*^\top \boldsymbol{v}_*)^2\vert$ is the alignment between $\boldsymbol{v}_{t-1}^*$ and $\boldsymbol{v}_*$, and $\epsilon=\min_{\boldsymbol{U}} \Vert\boldsymbol{B}_{t-1}^*-\boldsymbol{U}\boldsymbol{B}_*\Vert^2_2$ is the distances between $\boldsymbol{B}_{t-1}^*$ and $\boldsymbol{B}_*$ under a rotation.

\end{theorem}
\begin{proof}
This theorem is an immediate deduction of Theorem 3.1 in \cite{lpft} by simply regarding inputs of task $t$ as in-distribution data, and inputs of task $t'$ as out-of-distribution data.
\end{proof}
As for multi-head setting, we have 
\begin{align*}
&\sqrt{\mathcal{L}_{t'}(\boldsymbol{B}_{t}^*, \boldsymbol{v}_{t'}^*)}=\sqrt{\max_{\Vert \boldsymbol{x}\Vert\leq 1}( \boldsymbol{v}_{t'}^{*\top} \boldsymbol{B}_{t}^*\boldsymbol{x} - \boldsymbol{v}_{t'}^{*\top} \boldsymbol{B}_{t'}^*\boldsymbol{x})^2}\\
&=\Vert\boldsymbol{v}_{t'}^{*\top} \boldsymbol{B}_{t}^*-\boldsymbol{v}_{t'}^{*\top} \boldsymbol{B}_{t'}^*\Vert_2\\
&\geq\Vert\boldsymbol{v}_{t}^{*\top} \boldsymbol{B}_{t}^*-\boldsymbol{v}_{t'}^{*\top} \boldsymbol{B}_{t'}^*\Vert_2-\Vert\boldsymbol{v}_{t}^{*\top} \boldsymbol{B}_{t}^*-\boldsymbol{v}_{t'}^{*\top} \boldsymbol{B}_{t}^*\Vert_2\\
&=\sqrt{\mathcal{L}_{t'}(\boldsymbol{B}_{t}^*, \boldsymbol{v}_{t}^*)}-\epsilon_2
\end{align*}
in which $\epsilon_2=\Vert\boldsymbol{v}_{t}^{*\top} \boldsymbol{B}_{t}^*-\boldsymbol{v}_{t'}^{*\top} \boldsymbol{B}_{t}^*\Vert_2$ is the divergence between the two sub-networks corresponding to the two heads. We can regard $\epsilon_2$ as a relaxation provided by multi-head setting.

\subsection{Proof of Theorem 3}
\begin{theorem}
If for all task $t'\leq t$: (i) $\boldsymbol{v}$ is initialized with $\boldsymbol{v}_{t'}^{lp}$, and (ii) there exists $\boldsymbol{v}_0$ such that $\mathcal{L}_{t'}(\boldsymbol{B}_{0}, \boldsymbol{v}_0)=\mathcal{L}_{t'}(\boldsymbol{B}_{t'}^*, \boldsymbol{v}_{t'}^*)$ (i.e. $\boldsymbol{B}_{0}$ is good enough), then for all tasks $t'\leq t$:
$$\mathcal{L}_{t'}(\boldsymbol{B}_{t}^{*}, \boldsymbol{v}_{t'}^*)=\mathcal{L}_{t'}(\boldsymbol{B}_{t'}^{*}, \boldsymbol{v}_{t'}^*)$$
\end{theorem}
\begin{proof}
For all task  $t'\leq t$, if $\boldsymbol{B}_{t'-1}^*=\boldsymbol{B}_{0}$  there exists $\boldsymbol{v}_0$, such that $\mathcal{L}_{t'}(\boldsymbol{B}_{t'-1}^*, \boldsymbol{v}_0)=\mathcal{L}_{t'}(\boldsymbol{B}_{t'}^*, \boldsymbol{v}_{t'}^*)$, than we can know form Proposition 3.2 in \cite{lpft} that $\boldsymbol{v}_{t'}^{lp}=\boldsymbol{v}_0$. Since overparameterized model can achieve zero loss, we have $\mathcal{L}_{t'}(\boldsymbol{B}_{t'-1}^*, \boldsymbol{v}_{t'}^{lp})=\mathcal{L}_{t'}(\boldsymbol{B}_{0}, \boldsymbol{v}_0)=0$. So the gradient is also zero, and thus the feature extractor does not change at all, which means $\boldsymbol{B}_{t'}^*=\boldsymbol{B}_{0}$. Then inductively, for all task  $t'\leq t$, we have $\boldsymbol{B}_{t'}^*=\boldsymbol{B}_{0}$, which leads to $\mathcal{L}_{t'}(\boldsymbol{B}_{t}^{*}, \boldsymbol{v}_{t'}^*)=\mathcal{L}_{t'}(\boldsymbol{B}_{t'}^{*}, \boldsymbol{v}_{t'}^*)$ as we desire.
\end{proof}

\begin{table*}
\centering
\scalebox{0.9}{
\begin{tabular}{p{1cm}<{\centering}p{2cm}<{\centering}p{7cm}<{\centering}p{1.5cm}<{\centering}p{1.5cm}<{\centering}p{1.5cm}<{\centering}p{1.5cm}<{\centering}}
\toprule[1.0pt]
\textbf{Method}&Hyperparameter&Search Range  & CIFAR100&CUB200&Caltech101&Flowers102  \\
\specialrule{0em}{1pt}{1pt}

\specialrule{1pt}{1pt}{1pt}

\specialrule{0em}{1pt}{1pt}

EWC      & $\lambda$  & \{3e2, 1e3, 3e3, 1e4, 3e4, 1e5, 3e5\}   &1e3 &3e3&3e4 &1e5  \\\specialrule{0em}{1pt}{1pt}
SI    & $c$ & \{1, 2, 5, 10, 20, 50\}   &2 &5&10 &20 \\\specialrule{0em}{1pt}{1pt}
RWalk    & $\lambda$ & \{5, 10, 20, 50, 100, 200\}    &20 &20&50 &100   \\\specialrule{0em}{1pt}{1pt}
MAS &$\lambda$& \{1, 3, 10, 30, 100, 300\}   &10 &30&30 &30  \\\specialrule{0em}{1pt}{1pt}

CPR   & $\beta$ & \{0.005, 0.01, 0.02, 0.05, 0.1, 0.2, 0.5\}   &0.2 &0.02&0.01 &0.1    \\\specialrule{0em}{1pt}{1pt}
AFEC  & $\lambda_e$ & \{0.001, 0.01, 0.1, 1, 10, 100, 1000, 10000\}   &0.01 &100&1000 &10 \\\specialrule{0em}{1pt}{1pt}

\bottomrule[1.0pt]

\end{tabular}
}
\caption{Results of hyperparameter search.}
\label{app_table2}
\end{table*}

\section{Experiment Details}
\subsection{Dataset}
For CIFAR100 and CUB200, we use the official train/test split. For Caltech101 and Flowers102, we randomly choose 80\% images as training set, and the others as test set. During hyperparameters search, we further randomly choose 10\% images in training set for validation. The details of dataset statistics are in Table \ref{app_table1}.
\subsection{Implementation}
All experiments are implemented  using \emph{Pytorch} 1.10 with CUDA 10.2 on NVIDIA 1080Ti GPU. The datasets and dataloaders are conducted via \emph{Avalanche} \cite{Avalanche}. We use ResNet18 implemented by \emph{timm} \cite{timm}, and its pre-trained parameters are downloaded from \emph{torchvision}. The baselines are reproduced on the basis of the official codes of \cite{cpr,afec,ccll}, in which we only modify the dataset interfaces. 
\subsection{Data Pre-processing}
For each image, we pre-process it as follows (Pytorch style):
\begin{lstlisting}
Compose(
    Resize(size=256, interpolation=bilinear, max_size=None, antialias=None)
    CenterCrop(size=(224, 224))
    ToTensor()
    Normalize(mean=[0.4850, 0.4560, 0.4060], std=[0.2290, 0.2240, 0.2250])
)
\end{lstlisting}
There is no extra data augmentation applied.
\subsection{Hyperparameters Search}
For each baseline, we extensively search the best hyperparameter, which is shown in Table \ref{app_table2}. We train the model for fixed epochs for each method, which depends on the complexity of the dataset. We randomly choose 10\% images in training set for validation, exclude them from training, and use them to determine the best hyperparameters. After the hyperparameter search, we retrain the model with the whole training set and report the results on test set. Note that our ConFiT does not need hyperparameter search.

\section{Difference between IRS and \emph{Internal Covariate Shift}}
We notice that there is another concept called Internal Covariate Shift (ICS) about BN. To avoid confusion, we here clarify the difference between IRS and ICS. 

Firstly, we would emphasize that ICS and IRS are totally different concepts. ICS describes the instability of intermediate representations' distribution during the training stage, i.e. the inputs of intermediate layers are unstable \textbf{in training}, which impedes the  learning of networks. One of the motivations of BN is to address ICS, so as to make the network easier to train.

Whereas IRS is defined in the scenario of continual learning, which means the intermediate representations of \textbf{data of previous task} were shifted, because the network is fitting the data of the newest task. The shifted representation will disrupt the function of BN \textbf{in testing}, since the running moments will  no longer be representative of the true moments of intermediate representations.

Overall, BN can solve ICS, but will suffer from IRS in continual learning, which is what we have attempted to address.

\bibliographystyle{named}
\small{\bibliography{ijcai22}}